\documentclass[conference, compsoc, onecolumn, 11pt, letter]{IEEEtran}

\usepackage{cite}
\usepackage{graphicx}
\usepackage{epsfig}
\usepackage[cmex10]{amsmath}
\usepackage{multirow}

\newcommand{\squishlist}{
 \begin{list}{$\bullet$}
  { \setlength{\itemsep}{0pt}
     \setlength{\parsep}{3pt}
     \setlength{\topsep}{3pt}
     \setlength{\partopsep}{0pt}
     \setlength{\leftmargin}{1.5em}
     \setlength{\labelwidth}{1em}
     \setlength{\labelsep}{0.5em} }}

\newcommand{\squishlisttwo}{
 \begin{list}{$\bullet$}
  { \setlength{\itemsep}{0pt}
     \setlength{\parsep}{0pt}
    \setlength{\topsep}{0pt}
    \setlength{\partopsep}{0pt}
    \setlength{\leftmargin}{2em}
    \setlength{\labelwidth}{1.5em}
    \setlength{\labelsep}{0.5em}}}  

\newcommand{\squishend}{
  \end{list}  }

\begin{document}
\title{Compression of Deep Neural Networks for Image Instance Retrieval}

\author{
\IEEEauthorblockN{Vijay Chandrasekhar$^{1,4}$, Jie Lin$^{1}$, Qianli Liao$^{3}$, \\ Olivier Mor\`ere$^{2}$, Antoine Veillard$^{2}$, Lingyu Duan$^{5}$, Tomaso Poggio$^{3}$\\ 
}
\IEEEauthorblockA{
%$^1$Information System Labs, Stanford University, CA\\
$^1$Institute for Infocomm Research, Singapore\\
$^2$ Universit\'e Pierre et Marie Curie, Paris, France\\
%$^3$ Center for Brains, Minds and Machines, McGovern Institute, Boston, MA\\
$^3$ Massachusetts Institute of Technology, Boston, MA\\
$^4$ Nanyang Technological University, Singapore \\
$^5$ Peking University, China \\
}
}
\maketitle

\begin{abstract}

Image instance retrieval is the problem of retrieving images from a database which contain the same object.
Convolutional Neural Network (CNN) based descriptors are becoming the dominant approach for generating {\it global image descriptors} for the instance retrieval problem.
One major drawback of CNN-based {\it global descriptors} is that uncompressed deep neural network models require hundreds of megabytes of storage making them inconvenient to deploy in mobile applications or in custom hardware.
In this work, we study the problem of neural network model compression focusing on the image instance retrieval task.
We study quantization, coding, pruning and weight sharing techniques for reducing model size for the instance retrieval problem.
We provide extensive experimental results on the trade-off between retrieval performance and model size for different types of networks on several data sets providing the most comprehensive study on this topic. 
We compress models to the order of a few MBs: two orders of magnitude smaller than the uncompressed models while achieving negligible loss in retrieval performance.
\footnote{V. Chandrasekhar, L. Jie and Q. Liao contributed equally.}

\end{abstract}

\vspace{-0.1in}   
\section{Introduction}
\vspace{-0.1in}   

Image instance retrieval is the problem of retrieving images from a database representing the same object or scene as the one depicted in a query image.
The first step of a typical retrieval pipeline starts with the comparison of vectors representing the image content known as {\it global image descriptors}.
Deep neural networks have seen rapid progress in the last few years, and starting with their remarkable performance on the {\it ImageNet} large scale image classification challenge~\cite{AlexNet,Simonyan2014,ResNet}, they have become the dominant approach in a wide range of computer vision tasks.
In recent work~\cite{practicalguide2015,Yandex,sharif2015baseline,cbmmgroupinvariance}, Convolutional Neural Networks (CNN) have also been used to generate global feature descriptors for image instance retrieval, and are rapidly becoming the dominant approach for the retrieval problem.

While CNNs provide high performance, they suffer from one major drawback. 
State-of-the-art CNNs like AlexNet~\cite{AlexNet}, VGG~\cite{Simonyan2014} and Residual Networks~\cite{ResNet} consist of hundreds of millions of neurons.
Stored in floating point precision, these networks require hundreds of megabytes for storage.
Also, neural networks are getting deeper and deeper, as performance gains are obtained with increasing amounts of training data and increasing the number of layers: e.g., deep residual networks can be hundreds or even thousands of layers deep~\cite{ResNet}.

There are many practical reasons why smaller networks are desirable.
First, there are several applications, where image classification or retrieval needs to be performed on a mobile device, which require the CNN to be stored on the client.
Mobile applications which are hundreds of megabytes in size are not practical.
Second, as networks get larger, it is not feasible to train them on a single machine. 
Large neural networks are trained across multiple machines, and one of the key bottlenecks in training is the neural network weights or gradient data that are transferred between machines in the distributed gradient descent optimization step.
Better weight compression will make training larger networks more practical.
Third, there is immediate need for smaller networks for efficient hardware implementations of deep neural networks.
Storing the entire network on chip will allow fast access, reduce processing latency and improve energy efficiency.
Fourth, emerging MPEG standards like Compact Descriptors for Visual Search (CDVS)~\cite{CDVSOverview} and Compact Descriptors for Video Analysis (CDVA) \footnote{The MPEG CDVS/CDVA evaluation framework including test dataset is available upon request at http://www.cldatlas.com/cdva/dataset.html} require feature extraction to be performed with a few MB of memory to enable efficient implementations of streaming hardware. 
Without model compression, deep learning based descriptors cannot be adopted in these emerging standards.
Most of the recent work on model compression has been focused on compressing models for the image classification task: identifying the category that a query image belongs to.
The two tasks: image classification and instance retrieval, while related, pose different requirements.
CNNs consist of alternating convolutional and downsampling layers, and finally a set of one or more fully connected layers that map to a set of output classes or labels.
For image classification tasks, the fully connected layers serve as the classifier, mapping rich feature representations to output classes.
On the other hand, for image retrieval tasks, intermediate layers of the CNN have been shown to be effective high dimensional {\it global descriptors}~\cite{practicalguide2015}.

We highlight some of the recent work on model compression, where the primary focus has been on reducing model size while maintaining high image classification accuracy.
For architectures like~\cite{AlexNet}, the fully connected layers contain the highest number of parameters.
As a result, in~\cite{deepCompressionVQ}, Gong et al. propose Vector Quantization techniques for parameters in the fully connected layers, while leaving the convolutional layers untouched.
In their recent work, Han et al.~\cite{DeepCompression,han2015learning} propose quantization and coding techniques for compressing neural networks.
Han et al. prune the network by learning only the important connections required for the classification task. 
Following pruning, scalar quantization and huffman coding are used to further reduce model size.
In~\cite{SqueezeNet}, the authors propose a new architecture called {\it SqueezeNet}, which contains 50$\times$ fewer parameters than {\it AlexNet}~\cite{AlexNet}, while achieving similar classification performance.
The number of parameters in the network is reduced by smart choice of filter sizes and number of filters at each layer.
6-bit quantization is used to further reduce model size.

In this work, we propose quantization, pruning and coding techniques for compression of deep networks, with a focus on the image instance retrieval task, unlike~\cite{SqueezeNet,DeepCompression,deepCompressionVQ}.
We also study the problem in the context of state-of-the-art deep residual networks.
For residual networks, we propose sharing parameters across different layers to reduce the number of weights. 
Quantization and coding techniques specific to residual networks are further applied to reduce model size.
We perform extensive evaluation of trade-off between retrieval performance and model size for different types of networks on several data sets, providing the most comprehensive study on this topic. 
We compress models to the order of a few MBs: two orders of magnitude smaller than uncompressed deep networks, while achieving negligible loss in retrieval accuracy.

The paper is organized as follows.
In Section~\ref{sec:image_retrieval}, we discuss how CNNs are used for the image instance retrieval task.
Following that, in section~\ref{sec:model_compression}, we discuss quantization, coding, pruning and weight sharing techniques for state-of-the-art deep networks.
In Section~\ref{sec:exp}, we provide detailed experimental results and analysis of proposed methods.

\vspace{-0.1in}
\section{Image Retrieval with Deep Networks}
\vspace{-0.1in}

\label{sec:image_retrieval}

A typical image instance retrieval system starts with the comparison of vectors representing the image content, known as {\it global image descriptors}.

There is a growing body of work focused on using activations directly extracted from CNNs as {\it global descriptors} for image instance retrieval.
All popular CNN architectures share a set of common building blocks: a succession of convolution-downsampling operations designed to model increasingly high-level visual representations of the data.
Table~\ref{tab:arch} shows the model architecture for different networks considered in this work.

Initial studies~\cite{Yandex,Razavian2014} proposed using representations extracted from fully connected layers of CNNs as a global descriptor for image retrieval.
Promising results over traditional handcrafted descriptors like Fisher Vectors based on local SIFT features, were reported first in~\cite{Yandex,Razavian2014,practicalguide2015}.
Recent papers~\cite{Razavian15,sharif2015baseline,babenko2} show that spatial max and average pooling of feature maps output by intermediate convolutional layers is an effective representation, and higher performance can be achieved compared to using fully connected layers. 
 
Proposed techniques in~\cite{Razavian15,sharif2015baseline,babenko2} provide limited invariance to translation, but not to scale or rotation changes. 
To alleviate the scale issue, Tolias et al.~\cite{ToliasSJ15} proposed averaging of max pooled features over a set of multi-scale region of interest (ROI) feature maps in the image, similar to the R-CNN approach~\cite{girshick2014rcnn} used for object detection. 
Inspired from a recently proposed invariance theory for information processing~\cite{itheory1}, we proposed a Nested Invariance Pooling (NIP) method to produce compact and performant descriptors, invariant to translation, scale and rotation~\cite{cbmmgroupinvariance}. 

\begin{figure}
\centering{
\includegraphics[width=4in]{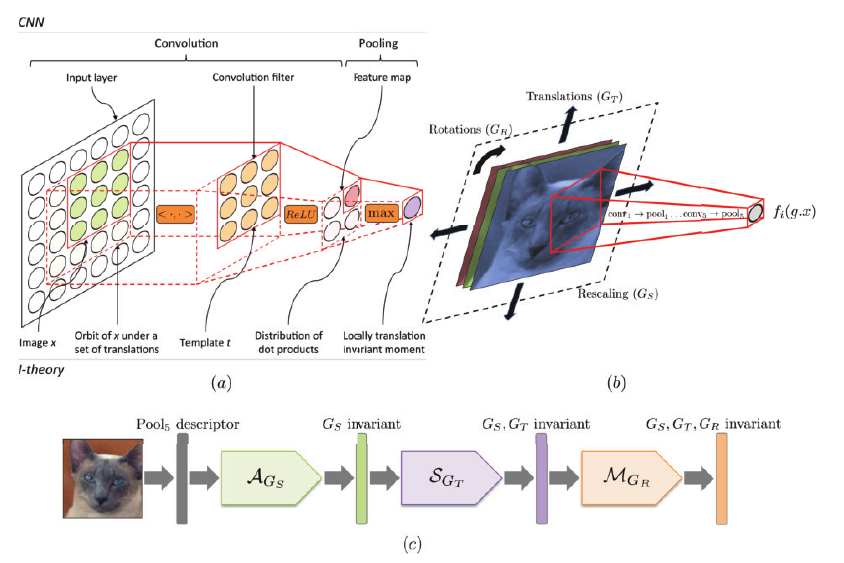}
\caption{\small (a) A single convolution-pooling operation from a CNN schematized for a single input layer and single output neuron.
(b) A specific succession of convolution and pooling operations learnt by the CNN (depicted in red) computes the \emph{} feature, the last convolution layer, $f_i$ for each feature map $i$ from the RGB image data.
A number of transformations $g$ are applied to the input $x$ in order to vary the response $f_i(g.x)$.
(c)
Starting with raw \emph{pool5} descriptors, an arbitrary number of transformation group invariances are stacked up.
$G_S,G_T,G_R$ corresponds to transformation groups scale, translation and rotation respectively, while $A_{G_S},S_{G_T},M_{G_R}$ refers to statistical moments average, standard deviation and max respectively, which are computed in a nested fashion.
}
}
\label{fig:invariance_method}
\end{figure}

In Figure~\ref{fig:invariance_method}, we provide a brief overview of NIP descriptors, which form the basis for all image retrieval experiments in this work.
Figure~\ref{fig:invariance_method}(a) shows a single convolution-pooling operation for a single input layer and single output neuron.
Figure~\ref{fig:invariance_method}(b) shows how a succession of convolution and pooling layers results in a set of feature maps $f_i$. 
A number of scale and rotation transformations are applied to the input image to obtain a series of feature maps.
In Figure~\ref{fig:invariance_method}(c), we show how feature maps are pooled in a nested fashion by computing statistical moments at each step (average, standard deviation, max).
The particular sequence of transformation groups and statistical moments are provided in~\cite{cbmmgroupinvariance,nipdeephash}.
Key to achieving high performance is stacking multiple transformation groups in a nested fashion, and pooling with increasing orders of moments.
Detailed evaluation provided in~\cite{nipdeephash} shows that NIP is robust to scale and rotation changes, and significantly outperforms other CNN based descriptors.
Next, we discuss compression of models shown in Table~\ref{tab:arch}.

\newcommand{\block}[2]{\multirow{2}{*}{\(\left[\begin{array}{c}\text{3$\times$3, #1}\\[-.1em] \text{3$\times$3, #1} \end{array}\right]\)}
}
\newcommand{\blocka}[2]{\multirow{3}{*}{\(\left[\begin{array}{c}\text{3$\times$3, #1}\\[-.1em] \text{3$\times$3, #1} \end{array}\right]\)}
}
\newcommand{\blockb}[3]{\multirow{3}{*}{\(\left[\begin{array}{c}\text{1$\times$1, #2}\\[-.1em] \text{3$\times$3, #2}\\[-.1em] \text{1$\times$1, #1}\end{array}\right]\)$\times$#3}
}
\newcommand{\blockc}[3]{\multirow{3}{*}{\(\left[\begin{array}{c}\text{3$\times$3, #2}\\[-.1em] \text{3$\times$3, #2}\\[-.1em] \text{1$\times$1, #1}\end{array}\right]\)}
}

\newcommand{\blockd}[2]{\multirow{2}{*}{\(\left[\begin{array}{c} \text{3$\times$3, #1}\\[-.1em] \text{3$\times$3, #1} \end{array}\right]\)$\times$#2}
}

\renewcommand\arraystretch{1.1}
\setlength{\tabcolsep}{3pt}
\begin{table*}[t]
\begin{center}
\resizebox{0.7\linewidth}{!}{
%\footnotesize
\begin{tabular}{c|c|c|c|c|c}
\hline
layer name & AlexNet & VGG-16 & ResNet50 & ResNet152 & Shared ResNet\\
\hline
\multirow{2}{*}{conv1} & \multirow{2}{*}{5$\times$5, 96} & \block{64}{1} & \multirow{2}{*}{7$\times$7, 64} & \multirow{2}{*}{7$\times$7, 64} & 
\multirow{2}{*}{7$\times$7, 64} \\
  &  &  &  &  &\\
\hline
\multirow{3}{*}{conv2\_x} & \multirow{3}{*}{3$\times$3, 256} &  \blocka{128}{1}  & \blockb{256}{64}{3} & \blockb{256}{64}{3} & 
\blockd{64}{2} \\
  &  &  &  & &\\
  &  &  &  & &\\
\hline
\multirow{3}{*}{conv3\_x} & \multirow{3}{*}{3$\times$3, 384} & \blockc{256}{256}{1}  & \blockb{512}{128}{4}  & \blockb{512}{128}{8} & 
\blockd{128}{3}\\
  &  &  &  & &\\
  &  &  &  & &\\
\hline
\multirow{3}{*}{conv4\_x} & \multirow{3}{*}{3$\times$3, 384}  & \blockc{512}{512}{1}  & \blockb{1024}{256}{6} & \blockb{1024}{256}{36} & 
\blockd{256}{10}\\
  &  &  &  & &\\
  &  &  &  & &\\
\hline
\multirow{3}{*}{conv5\_x} & \multirow{3}{*}{3$\times$3, 256}  & \blockc{512}{512}{1}  & \blockb{2048}{512}{3}  & \blockb{2048}{512}{3} & \blockd{512}{3}\\
  &  &  &  & &\\
  &  &  &  & &\\
\hline
\end{tabular}
}
\end{center}
\vspace{-.5em}
\caption{\footnotesize Architectures for different networks. Fully connected layers are discarded. Building blocks are shown in brackets, with the numbers of blocks stacked.
\vspace{-20pt}
}
\label{tab:arch}
\vspace{-.5em}
\end{table*}

\vspace{-0.1in}
\section{Model Compression}
\vspace{-0.1in}
\label{sec:model_compression}

State-of-the-art CNNs commonly consist of alternating convolutional and downsampling layers, and finally one or more fully connected layers that map to a set of output classes or labels.
Since we are interested in the task of instance retrieval and not image classification, fully-connected layers which produce inferior global descriptors can be discarded.
We consider the following class of networks in this work: {\it AlexNet}~\cite{AlexNet}, {\it VGG}~\cite{Simonyan2014}, 52-layer and 152-layer residual networks~\cite{ResNet} ({\it ResNet}), and residual networks with shared parameters ({\it Shared ResNet})~\cite{SharedResNet}.
Table~\ref{tab:arch} details the relevant part of the architecture of the different networks.

\begin{figure*}
	\centering{
		\begin{tabular}{@{}@{}c@{}@{} @{}@{}c@{}@{} @{}@{}c@{}@{} }

			\includegraphics[width=2in]{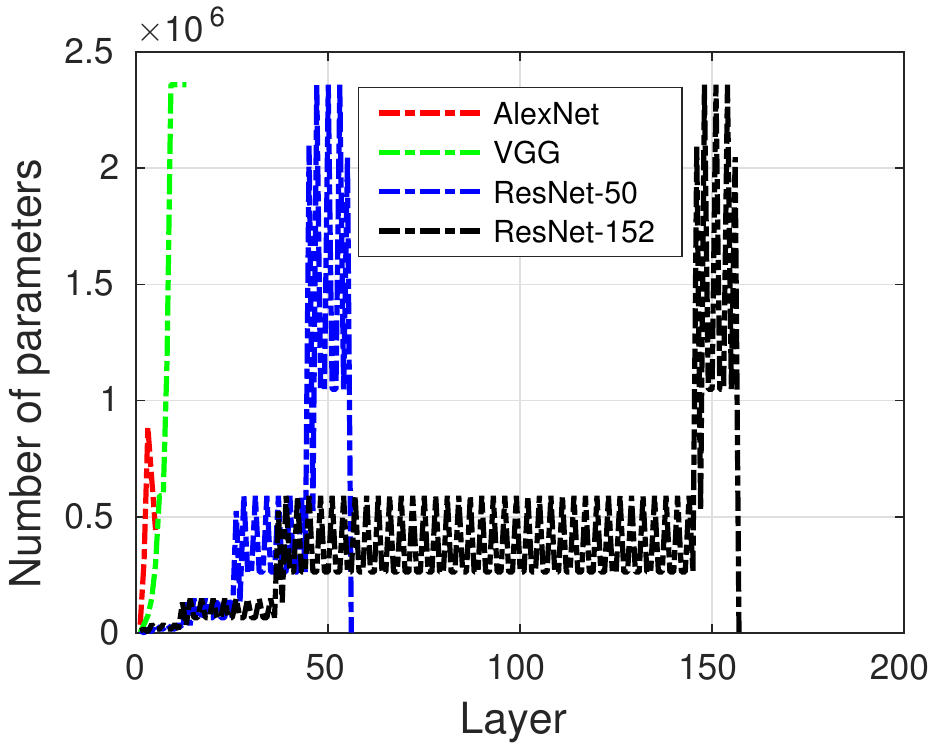} &
			\includegraphics[width=2in]{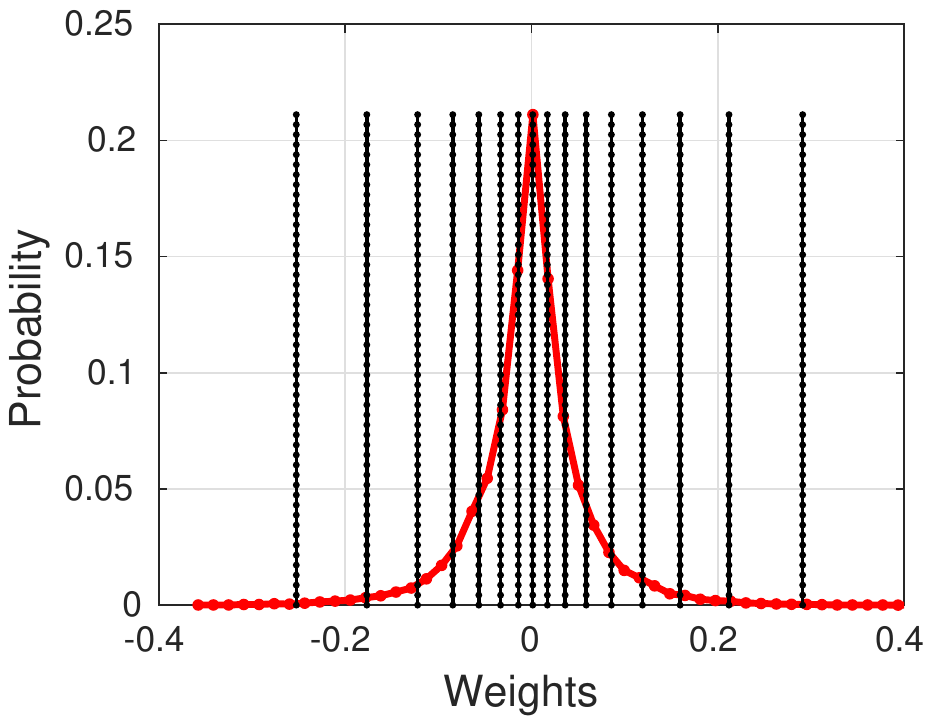} & 
			\includegraphics[width=2in]{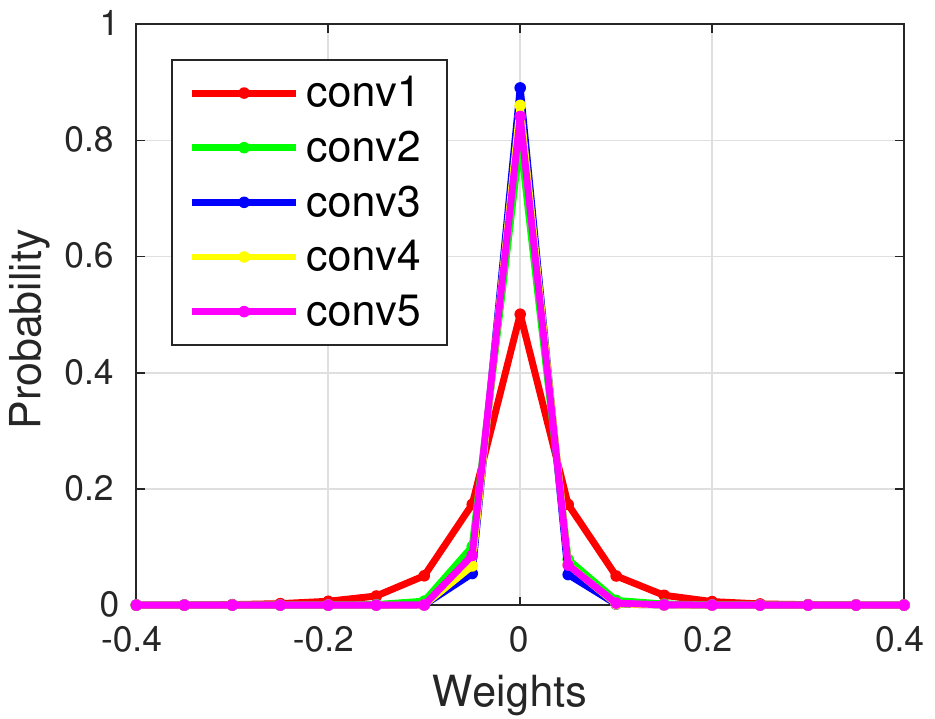} \\
			(a)  & (b) & (c) \\
		\end{tabular}
		\caption{{\footnotesize (a) Number of parameters in convolutional layers for different networks. (b) Distribution of weights for {\it conv1} layer of {\it AlexNet} and decision boundaries of Lloyd-Max quantizer. (c) Distribution of weights across different convolutional layers for {\it AlexNet}. 
			}}
			\label{fig:num_params}
		}	
\end{figure*}

Table~\ref{tab:model_params} lists the number of parameters in convolutional layers for each network. 
In Figure~\ref{fig:num_params}(a),  we plot the number of parameters for each convolutional layer. 
We note there are millions of parameters in the convolution layers, and the number of parameters typically increases with depth.
For {\it AlexNet} and {\it VGG}, the majority of parameters lie in the fully connected layers: discarding fully connected layers reduces the number of parameters in the network from 60M to 2.3M for {\it AlexNet}~\cite{AlexNet}.
For residual networks, the number of parameters in the convolutional layers far exceeds the number of parameters in the fully connected layers, as there are many more convolutional layers and less fully connected layers.
Even after discarding fully connected layers, uncompressed {\it VGG} and {\it ResNets} require more than 50 and 100 MB, motivating the need for compression.
Next, we discuss four building blocks for model compression.

\vspace{-0.1in}
\subsection{Quantization and Coding}
\vspace{-0.1in}
In Figure~\ref{fig:num_params}(b) and (c), we show the distribution of weights in different convolutional layers {\it conv1} to {\it conv5} for {\it AlexNet}. 
We note that the convolutional weight parameters follow a Laplacian distribution.
The distributions become more peaky (decreasing variance) with depth. 
This is intuitive as feature response maps become increasingly sparse with depth as higher level object representations are learnt. 
Similar trends are observed for other networks.

For each layer, we explore both scalar and vector quantization (VQ) techniques using the Lloyd-Max algorithm.
For simplicity, we ignore entropy in the centroid training step.
For VQ, we consider blocks of 2 and 4 with the number of codewords in the range of 256-1024. 
VQ additionally requires the codebook to be transmitted along with the network. 
With increasing codebook size, the size of codebook data becomes non-negligble compared to the weights.
In~\cite{deepCompressionVQ}, Gong et al. focus on compressing fully connected layers with VQ with large codebooks, which is less applicable for compressing convolutional layers.
Further, we explore applying different quantization parameters to different layers and study their impact on retrieval performance. 

We also explore how variable length coding with Huffman codes can be used to further reduce model size.
Variable length coding can reduce model size when models need to be transmitted over a network.
However, it is not feasible to uncompress the network each time feature extraction needs to be performed, in which case, fixed length coding is preferred.
Detailed experimental results are provided in Section~\ref{sec:exp}.

\vspace{-0.1in}
\subsection{Pruning}
\vspace{-0.1in}

One technique for reducing model size is more coarse quantization of weight parameters.
Another approach to trade-off model size for performance is to prune entire convolutional layers of the network, motivated by the visualization work in~\cite{zeiler}.
Zeiler and Fergus propose techniques for visualizing filter responses at different layers of a deep network~\cite{zeiler}.
The first convolutional layer typically learns Gabor-like filters, while the last layer represents high level concepts like cats and dogs.
As we go deeper into the network, the representations become more specific to object categories, while earlier layers provide more general feature representations for instance retrieval.
We reduce the number of parameters by varying the starting representation in Figure~\ref{fig:invariance_method}(c) for NIP based on earlier convolutional layers: {\it conv2} to {\it conv4}, instead of just the last layer before the fully connected layer: {\it pool5}.
Layers after the chosen convolutional layer are dropped from the network.
Note that this pruning approach is different from the pruning proposed in~\cite{DeepCompression} where connections are removed if they do not impact classification accuracy.

\vspace{-0.1in}
\subsection{Weight Sharing}
\vspace{-0.1in}

We propose sharing weights across layers to reduce the number of parameters in residual deep networks ({\it ResNets}). 
A {\it ResNet} adopts repeated residual blocks to facilitate learning ultra deep representations. 
A residual block typically consists of two or more convolutional layers and a identity shortcut mapping connecting its beginning and end.
A biological-inspired work~\cite{SharedResNet} showed that repeated residual blocks is isomorphic to a specific type of recurrent network (RNN) unrolled over time (see Figure~\ref{fig:all_about_resnets}(a)). 
Enforcing weight sharing (like RNN does) across repeated residual blocks preserves the performance of the corresponding {\it ResNet} while significantly lowers the number of parameters. 
In our experiments, we build on~\cite{SharedResNet} and repeat the residual blocks {\it conv1} to {\it conv4} 2, 3, 10 and 3 times respectively as illustrated in Figure~\ref{fig:all_about_resnets}(a). 
The resulting intermediate feature map sizes are also shown in Figure~\ref{fig:all_about_resnets}(a). 
These networks are trained on the ImageNet dataset, like the others. 
Quantization and coding are further applied to reduce model size.

\begin{table}
\centering
\begin{tabular}{|c|c|c|c|c|c|}
\hline
					& AlexNet 	& VGG 		& ResNet-52 	& ResNet-152 & Shared ResNet \\
\hline					
$\#$ of parameters	& 2.3M 	  	& 14.7M 	& 25.5M			& 60M 	& 8.4M \\
\hline
%\end{center}
\end{tabular}
\caption{\footnotesize Number of parameters (only convolutional) for each model}
\label{tab:model_params}
\vspace{-20pt}
\end{table}

\vspace{-0.1in}
\section{Experimental Results}
\vspace{-0.1in}
\label{sec:exp}

We study the trade-off between model size and performance on four popular instance retrieval data sets: {\it Holidays}, {\it Oxford buildings (Oxbuild)}, {\it UKBench (UKB)} and {\it Paris6k}.
Following standard protocol for {\it Holidays}, {\it Oxford5k} and {\it Paris6k}, retrieval performance is measured by Mean Average Precision (mAP).
For UKBench, we report the average number of true positives within the top 4 retrieved images (4$\times$Recall@4).
We start with off-the-shelf networks pre-trained on ImageNet classification data set.
When comparing pooling layers to fully connected layers on {\it AlexNet}, we resize all images to (227$\times$227) as fixed size input images are required. 
For all other experiments, if the longer side of input image exceeds 1024 pixels, we down sample the image to 1024, while maintaining aspect ratio.

\begin{figure*}
	\centering{
		\begin{tabular}{@{}@{}c@{}@{} @{}@{}c@{}@{\hskip 0.2in} @{}@{}c@{}@{} }

			\includegraphics[width=1.8in]{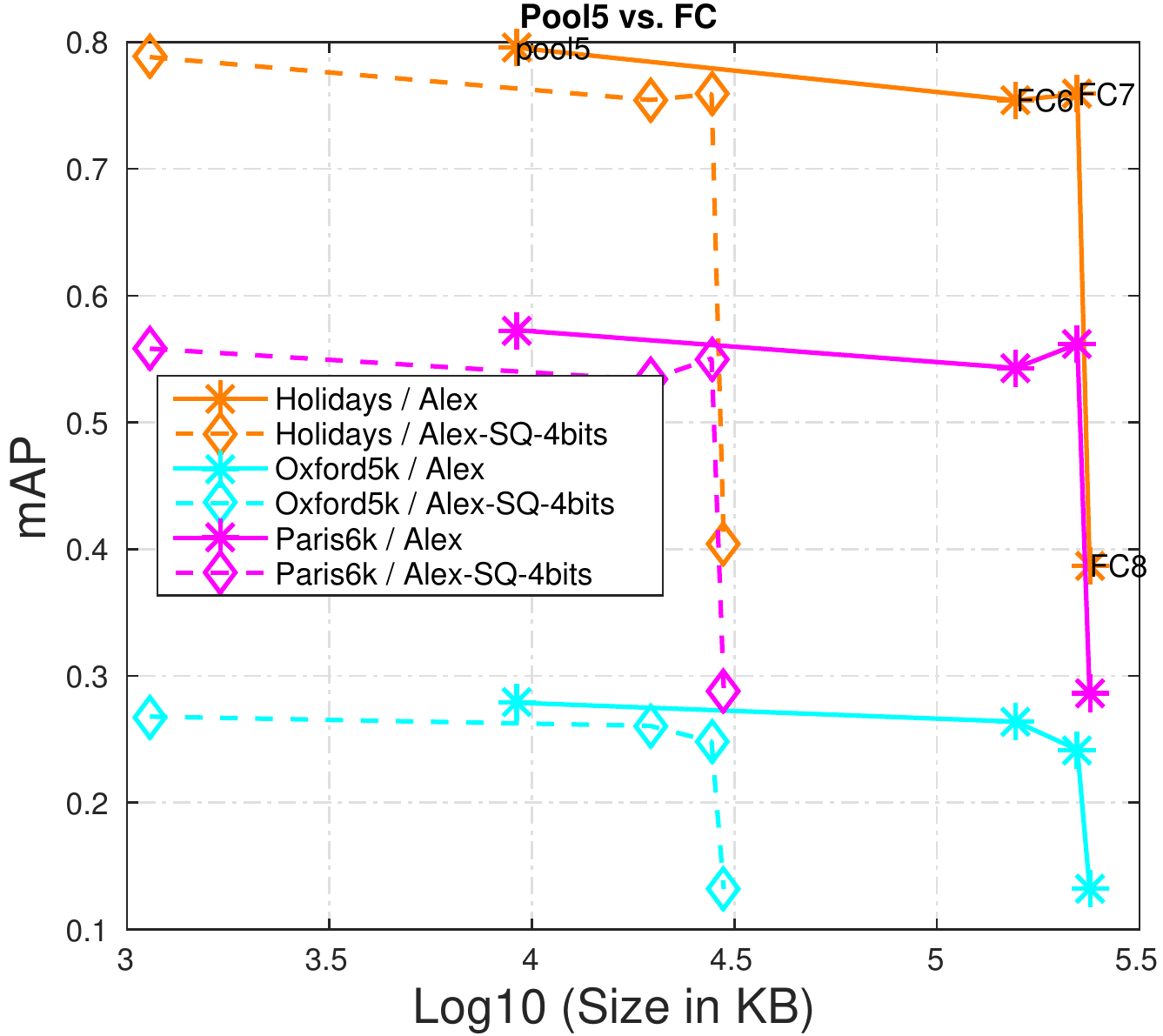} &
			\includegraphics[width=2in]{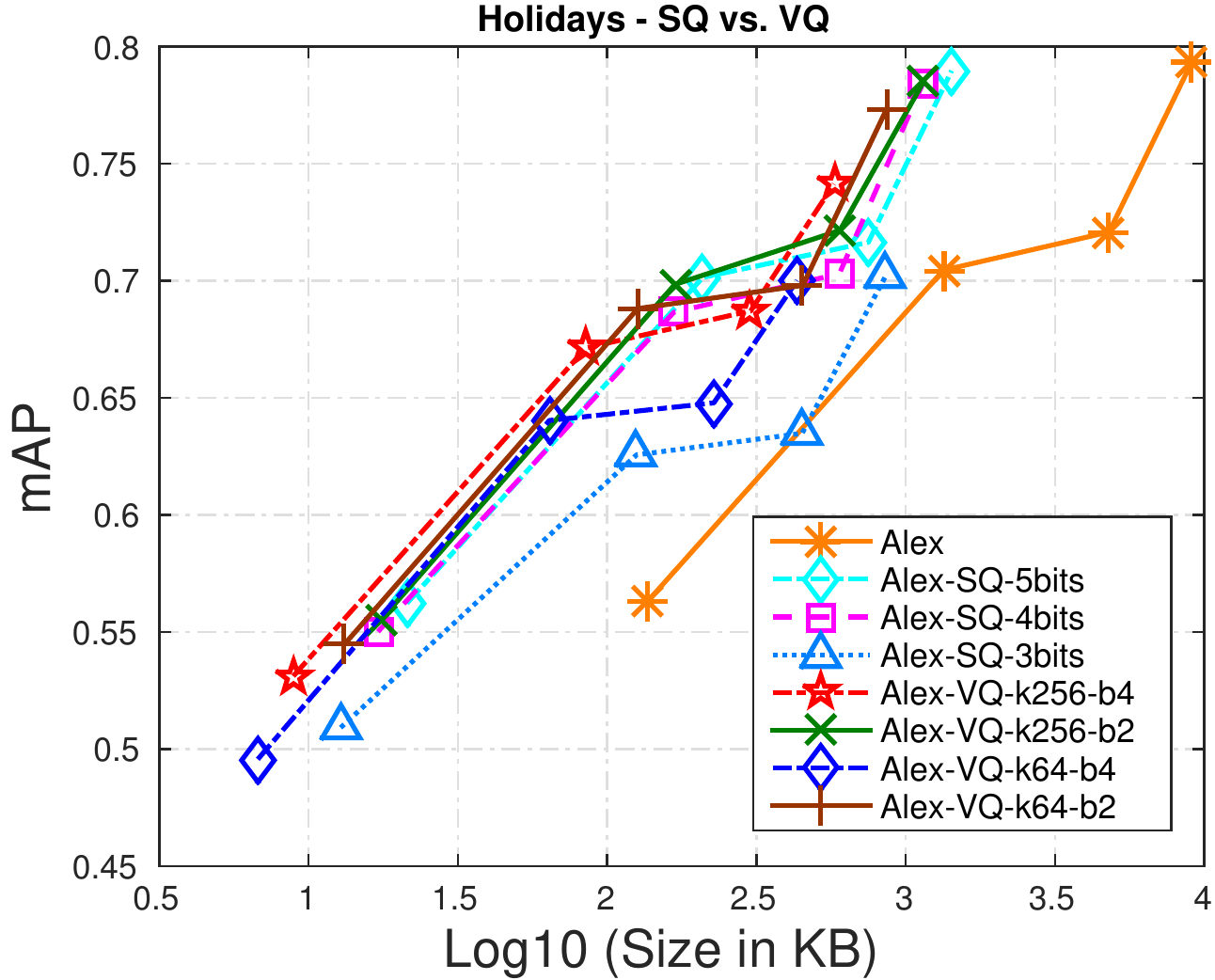} &
			\includegraphics[width=2in]{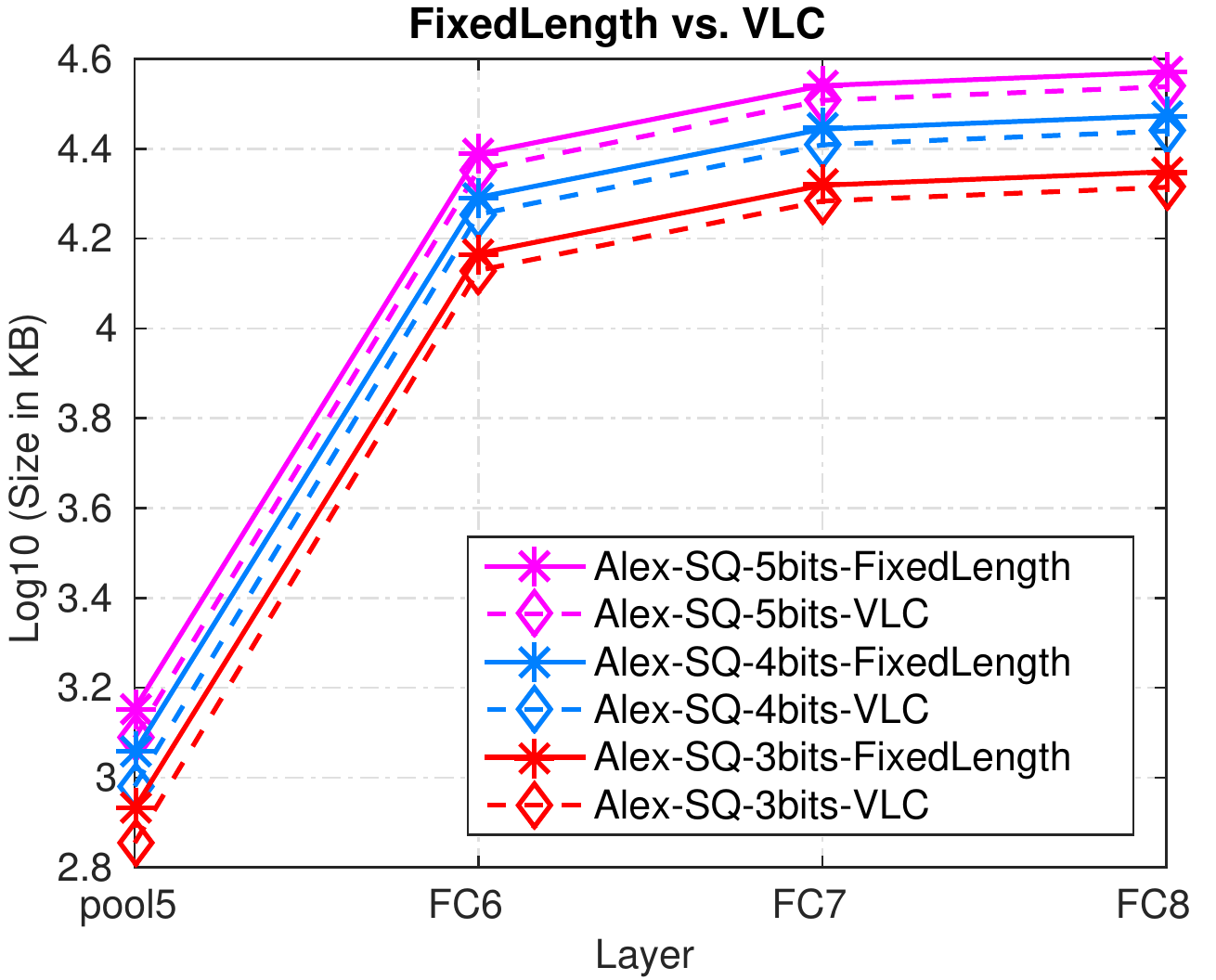} \\

		\end{tabular}
		\caption{{\footnotesize (a) Comparing {\it pool5} and fully connected layers on {\it AlexNet}. (b) Comparing SQ and VQ for varying parameters on {\it AlexNet}. (c) Comparing fixed length and variable length coding on {\it AlexNet}.
			}}
			\label{fig:general_results}
		}	
\end{figure*}

\begin{figure*}
	\centering{
		\begin{tabular}{@{}@{}c@{}@{} @{}@{}c@{}@{} }

			\includegraphics[width=2in]{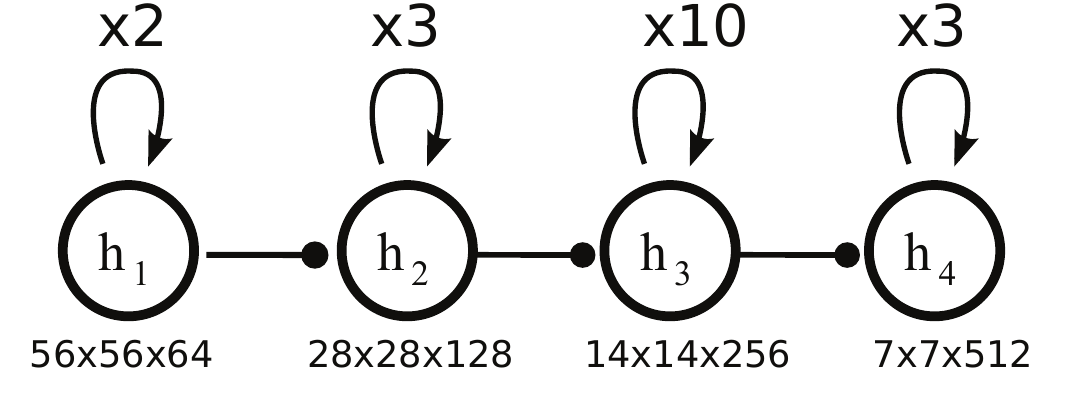} &
			\includegraphics[width=2in]{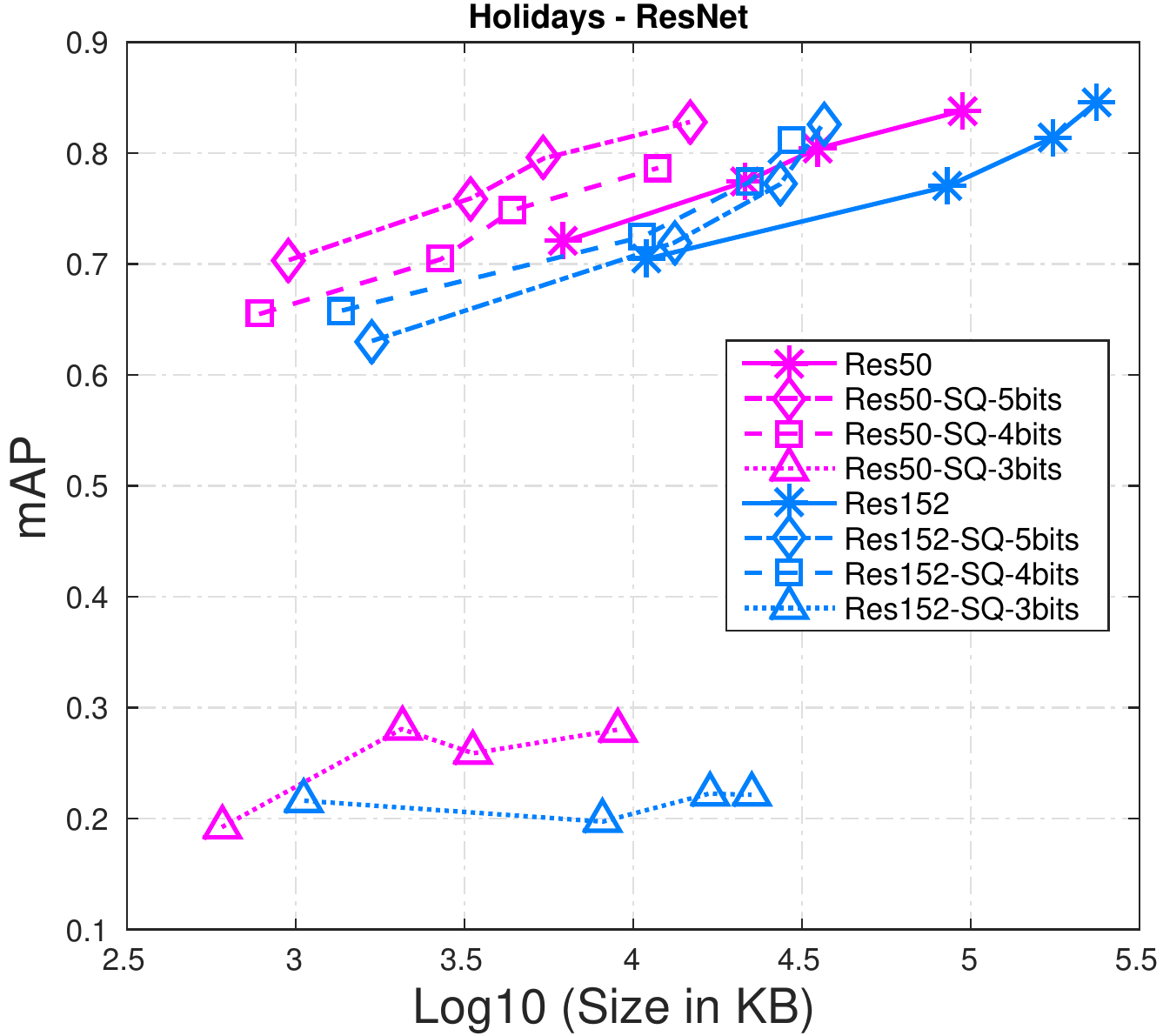} \\
			(a) & (b) \\
		\end{tabular}
		\caption{{\footnotesize  (a) Weight sharing across layers in residual networks. Building blocks in Table~\ref{tab:arch} {\it SharedNet} are repeated 2, 3, 10 and 3 times respectively with shared weights. (b) Effect of scalar quantization on residual networks. Retrieval performance drops steeply for 3-bit quantization.
			}}
			\label{fig:all_about_resnets}
		}	
\end{figure*}

\begin{figure*}
	\centering{
		\begin{tabular}{@{}@{}c@{}@{\hskip 0.5in} @{}@{}c@{}@{} }

			\multicolumn{2}{c}{\includegraphics[width=6in]{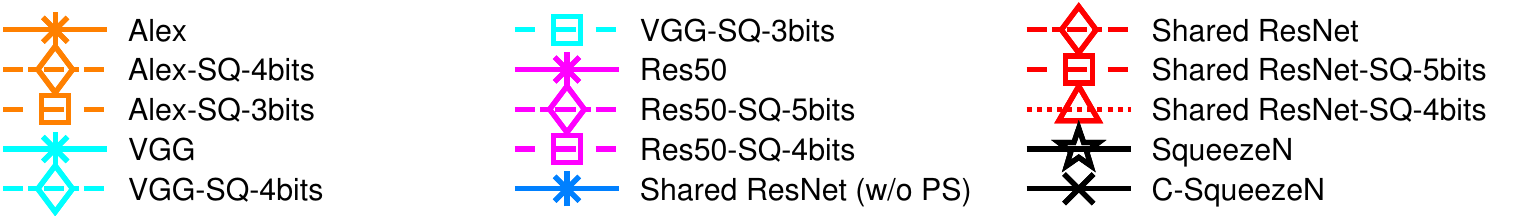}} \\
			\includegraphics[width=2.8in]{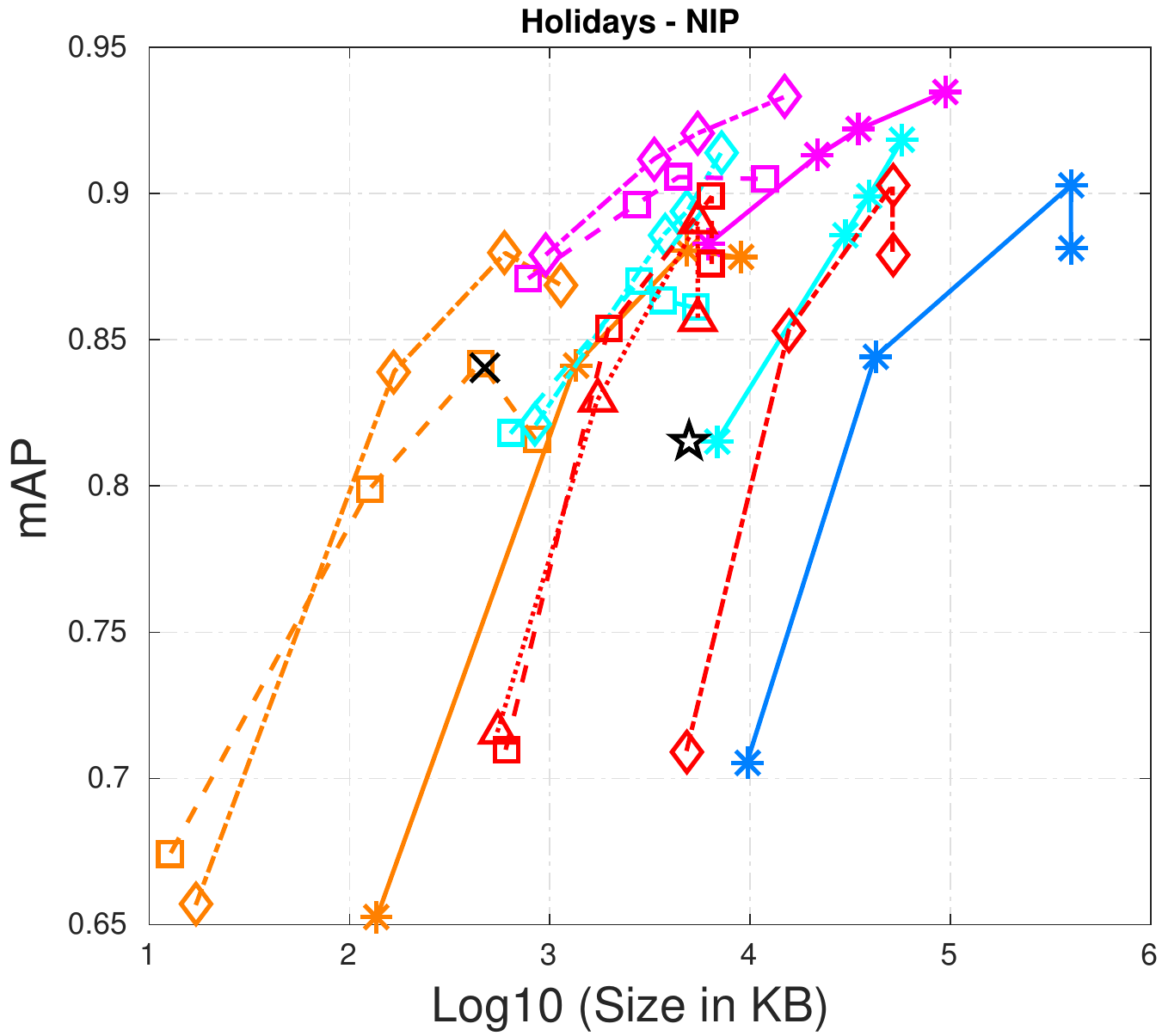} &
			\includegraphics[width=2.8in]{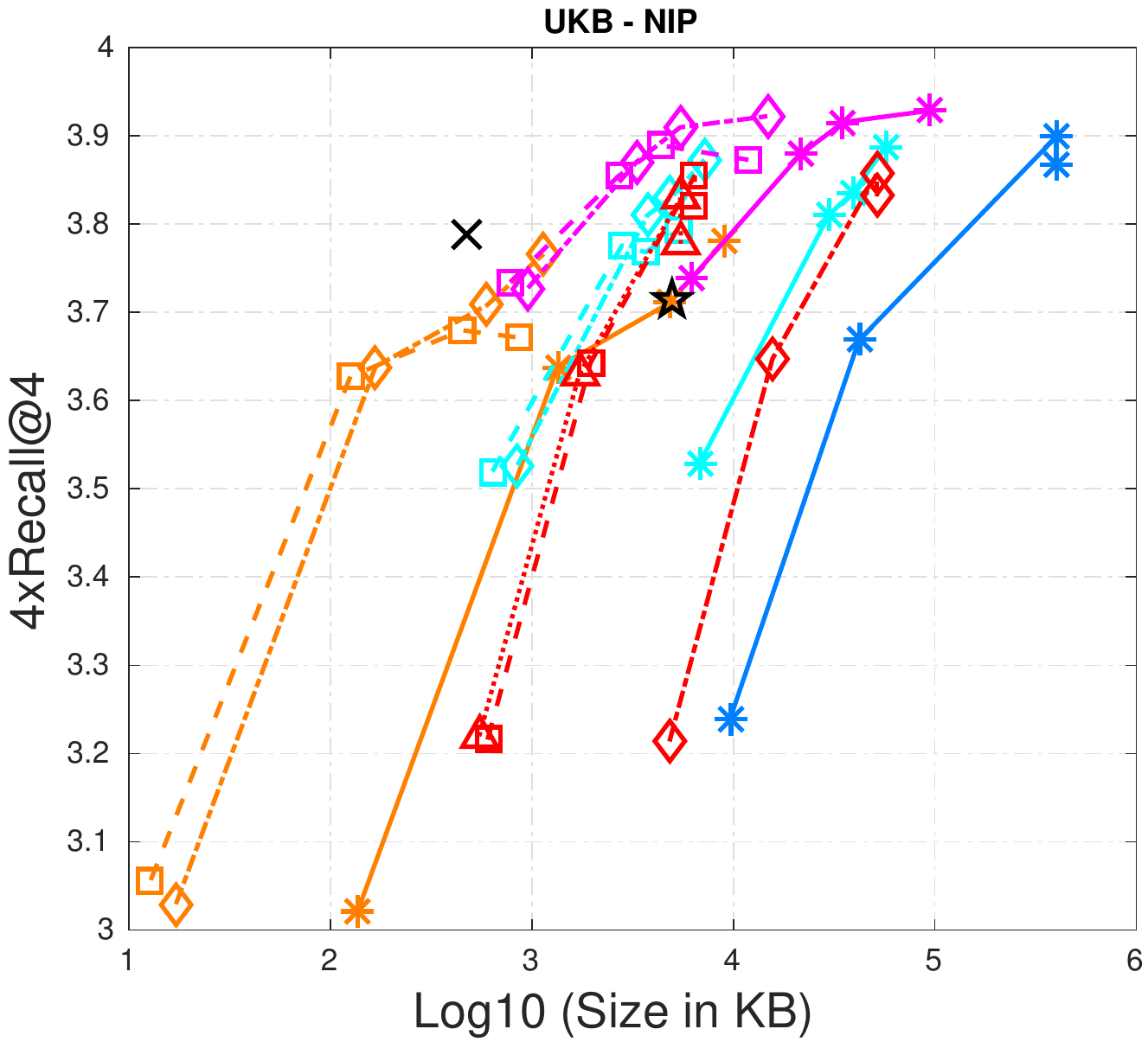} \\
			(a) Holidays & (b) UKB \\
			\\
			\includegraphics[width=2.8in]{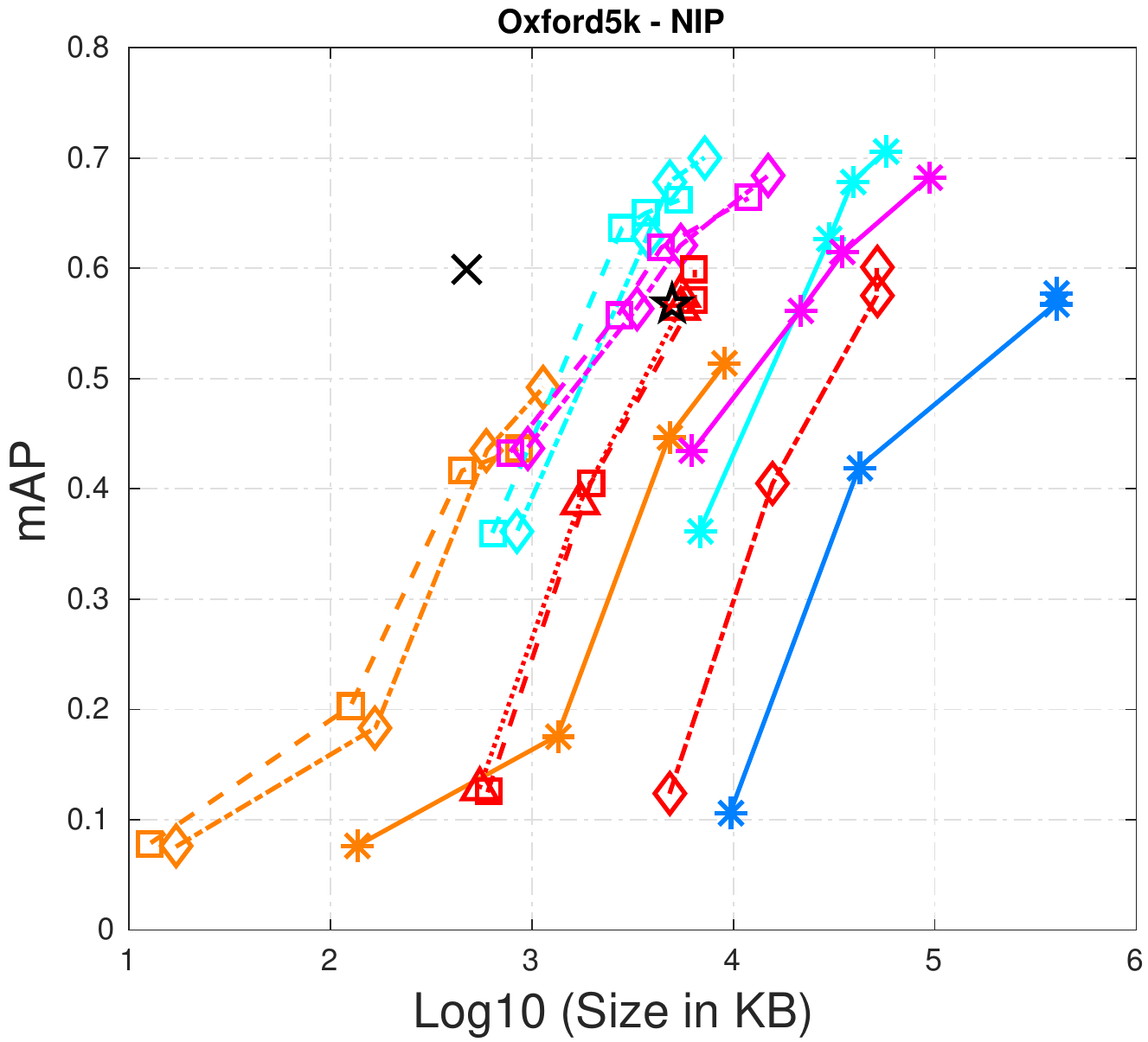} &
			\includegraphics[width=2.8in]{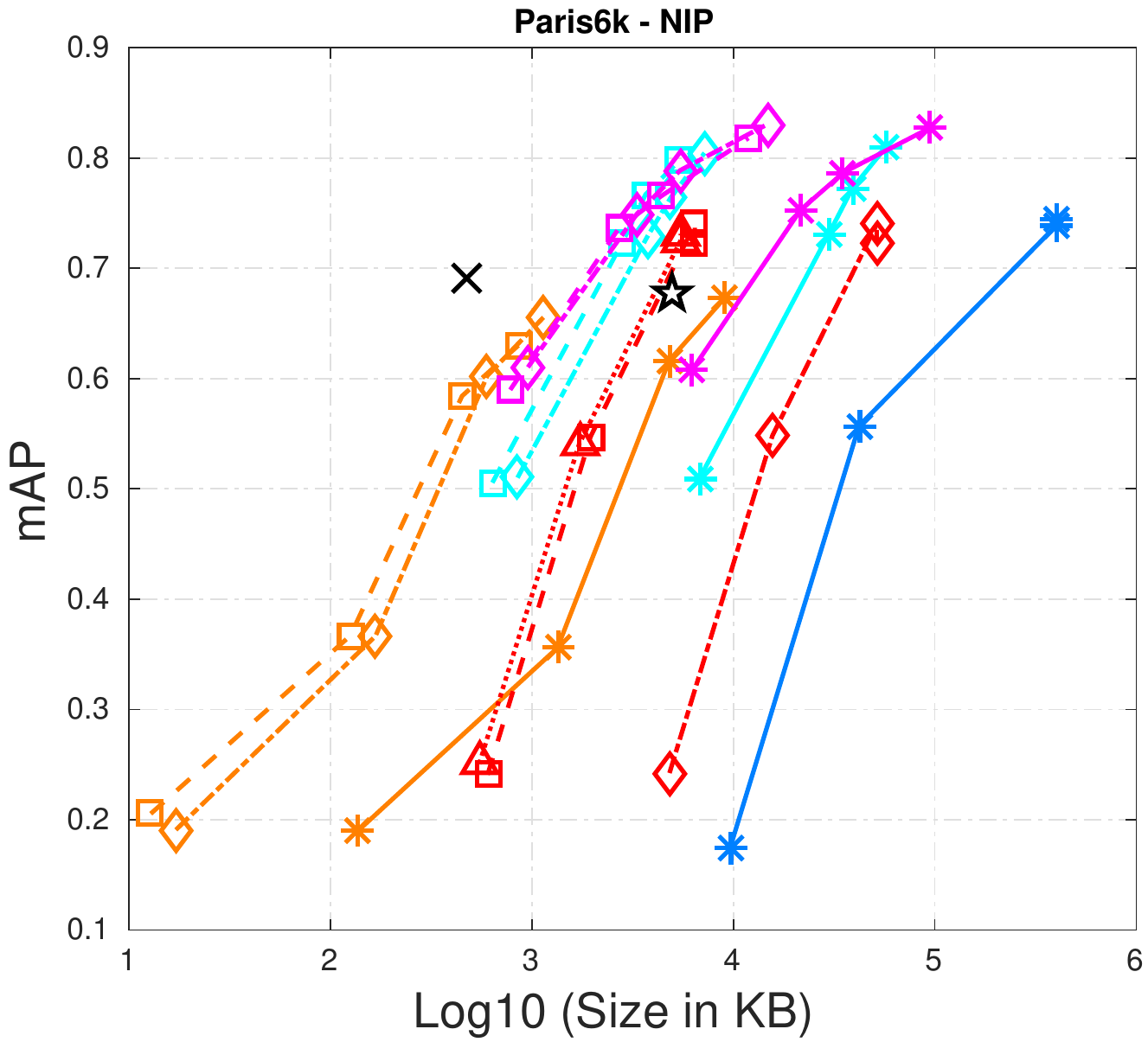} \\
			(a) Oxford-5K & (b) Paris-6K \\
		\end{tabular}
		\caption{{\footnotesize Comparing different model compression schemes on 4 different data sets. Points 3, 4 and 5 on the $x$ axis correspond to 1 MB, 10 MB and 100 MB respectively.  For each curve, model size is varied by pruning layers in the network.
			}}
			\label{fig:crazy_results}
		}	
\end{figure*}

To evaluate the performance of layer pruning, we choose 4 different layers for each network architecture.
In particular, {\it pool1}, {\it pool2}, {\it conv3\_relu}, {\it pool5} for {\it AlexNet}, {\it pool3}, {\it pool4}, {\it conv5a\_relu}, {\it pool5} for {\it VGG}, {\it res3d\_relu}, {\it res4c\_relu}, {\it res4f\_relu}, {\it pool5} for {\it ResNet-50} and {\it res3b7\_relu}, \\
{\it res4b15\_relu}, {\it res4b35\_relu}, {\it pool5} for {\it ResNet-152}.
For {\it Shared ResNet}, we use the 4 layers following convolutional layers {\it conv1} to {\it conv4} shown in Table~\ref{tab:arch}.
For NIP feature extraction, we use 4 rotations (0 to 360 degrees with step size 90 degrees), and for each rotated image, we sample 20 ROIs with 3 different scales.
We study the trade-off between model size and performance in Figures~\ref{fig:general_results},~\ref{fig:all_about_resnets} and \ref{fig:crazy_results}.
For each curve, the model size is varied by pruning each network back to intermediate layers as discussed above. 
Different curves represent compression with different quantization parameters.
We make several interesting observations.

\squishlist

\item We first compare the performance of the last convolutional layer {\it pool5} and full connected layers {\it fc6}, {\it fc7} and {\it fc8} for {\it AlexNet} on the {\it Holidays} data set in Figure~\ref{fig:general_results}(a).
Performance for {\it pool5} is higher than the fully connected layers. 
A significant drop is observed for {\it fc8}, which represents the layer corresponding to {\it ImageNet} class labels.
This confirms that fully connected layers can be discarded for the instance retrieval problem, while drastically reducing network size.
Similar trends are observed on other data sets, and confirms observations made in~\cite{babenko2}.

\item We compare Scalar Quantization (SQ) and Vector Quantization (VQ) for {\it AlexNet} on the {\it Holidays} data set in Figure~\ref{fig:general_results}(b). 
{\it k64-b2} for VQ corresponds to codebook size of 64 and block size of 2. 
We note that there is only a small gain with VQ compared to 4-bit or 5-bit SQ.  
This is intuitive, as one can expect weight parameters to be independent, and large codebooks for VQ are not feasible.
From here on, we use SQ in the rest of the experiments.

\item We compare Fixed Length Coding (FLC) and Variable Length Coding (VLC) for {\it AlexNet} on the {\it Holidays} data set in Figure~\ref{fig:general_results}(c). 
We observe a small but consistent gain of $\sim$15-20$\%$ with variable length coding, which can be useful when models need to be transmitted over a network.  
From here on, we use FLC in the rest of the experiments.

\item For compression of residual networks, we make two key observations.  First, we observe that performance drops drastically for 3-bit quantization for residual networks in Figure~\ref{fig:all_about_resnets}(b), compared to {\it AlexNet} in Figure~\ref{fig:general_results}(b)-(c). Residual networks are a lot deeper, and the quantization error accumulates over the layers leading to the steep drop. 
The residual networks in consideration also have batch normalization layers, which zero-center the data at every layer. 
Quantizing the Batch Normalization (BN) weights coarsely results in a drop in performance as shown in Table~\ref{tab:resnet_compression}.
As a result, we maintain the BN weights with floating point precision, while coarsely quantizing weight parameters in the rest of the experiments. 

\squishend

\begin{table}
\centering
\begin{tabular}{|c|c|c|c|}
\hline
Layer 	& 	Resnet-50  	& 	ResNet-50  & ResNet    \\
			&				&   4-bit Conv		  &	4-bit Conv			\\
			&				&   32-bit BN		  &	4-bit BN			\\
\hline
pool5 	& 0.8382		&	0.7806				& 0.7591			\\
\hline
%\end{center}
\end{tabular}
\caption{\footnotesize Coarse quantization of Batch Normalization (BN) parameters leads to drop in performance.}
\label{tab:resnet_compression}
\end{table}

We compare different model compression schemes in Figure~\ref{fig:crazy_results} and make the following salient observations.

\squishlist
\item Points 3,4,5 on the $x$ axis correspond to 1, 10 and 100 MB models respectively.
We observe that there is negligible loss in retrieval performance with 4-bit quantization for {\it AlexNet} and {\it VGG}.  
For residual networks, this is achieved with 5-bit quantization.   
2-bit quantization results in significant drop in performance for all models, and are not shown in the interest of space.
Compressed models in the order of 5-10 MB achieve performance almost similar to their uncompressed counterparts on each data set while being one to two orders of magnitude smaller.

\item Both pruning layers and coarser quantization of weight parameters provide similar trade-off in performance and model size.

\item Comparing compressed models, we note that compressed {\it VGG} achieves the highest performance on {\it Oxford5k}, while compressed {\it ResNet} achieves the highest performance on the other three data sets. Compressed {\it AlexNet} results in smaller networks, but performs significantly worse than top performing schemes.  

\item Compressed {\it SqueezeNet}~\cite{SqueezeNet} (C-SqueezeN) models are only 0.5 MB in size, and provide good trade-off in size and performance. However, they perform significantly worse than the best schemes. The aggressive optimizations used for reducing parameters in the {\it SqueezeNet} architecture hurt retrieval performance significantly.

\item For {\it Shared ResNet}, we first compare the models with and without parameter sharing, denoted as {\it Shared ResNet} and {\it Shared ResNet w/o PS}  respectively. Compared to no parameter sharing, 3-4 $\times$ smaller models are achieved by sharing weights. Further quantization of {\it Shared ResNet} results in smaller models.   One thing to note is that the {\it Shared ResNet} are trained from scratch on the {\it ImageNet} data set, and peak performance for our trained models is lower than the uncompressed {\it ResNet-50} of~\cite{ResNet}. Performance of compressed {\it Shared ResNet} can be further improved by starting with better residual network models. An alternative approach is to enforce shared weights across layers in the {\it ResNet} models of~\cite{ResNet} and fine-tune the network again.

\squishend

The experimental results above provide the most comprehensive study of model compression for the instance retrieval problem.
We provide several interesting directions for future work.
Models need to be made smaller further without losing retrieval performance: e.g., the Compact Descriptors for Visual Search (CDVS) standard only allows for 1 MB of memory for feature extraction in the low memory mode to enable streaming hardware implementations.  
Mathematical modelling of the distribution of weights would provide interesting insights for learning more compact models.
More sophisticated pruning techniques can be explored to reduce model size, with the explicit objective of maintaining high retrieval performance.
Performance of deep residual networks with shared weights needs to be improved further, and is a promising direction for achieving smaller performance models.

\vspace{-0.1in}
\section{Conclusion}
\vspace{-0.1in}
In this work, we studied the problem of neural network model compression focusing primarily on the image instance retrieval task.
We studied quantization, coding, pruning and weight sharing techniques for reducing model size for the instance retrieval problem.
Our compressed models are in the order of a few MBs: two orders of magnitude smaller than the uncompressed models while achieving negligible loss in retrieval accuracy.
We provide extensive experimental results on the trade-off between retrieval performance and model size for different types of networks on several data sets, providing the most comprehensive study on this topic. 

\section{Acknowledgement}
This work was supported in part by grants from National Natural Science Foundation of China (U1611461) and National Hightech Research and Development Program of China (2015AA016302).

\let\oldthebibliography=\thebibliography
  \let\endoldthebibliography=\endthebibliography
  \renewenvironment{thebibliography}[1]{%
    \begin{oldthebibliography}{#1}%
      \setlength{\parskip}{0ex}%
      \setlength{\itemsep}{0ex}%
  }%
  {%
    \end{oldthebibliography}%
  }
%\vspace{-0.05in}   

{
%\tiny{
\scriptsize{
\bibliography{marbib}   %>>>> bibliography data in report.bib
\bibliographystyle{IEEEtran}
}}

\end{document}